\pdfoutput=1
\documentclass[preprint,12pt]{elsarticle}

\usepackage{amssymb}
\usepackage{amsmath}
\usepackage{graphicx}

\usepackage{booktabs}
\usepackage{multirow}
\usepackage{siunitx}
\usepackage{tabularx}
\usepackage{threeparttable}
\usepackage{subcaption}
\usepackage{tikz}
\usepackage{float}
\usepackage{natbib}
\usetikzlibrary{shapes.geometric, arrows, positioning}
\journal{npj Digital Medicine}

\usepackage[hidelinks]{hyperref}

\begin{document}

\begin{frontmatter}

\title{Automated Multi-label Classification of Eleven Retinal Diseases: A Benchmark of Modern Architectures and a Meta-Ensemble on a Large Synthetic Dataset}

% Author Information
\author[aff1,aff2,aff3]{Jerry Cao-Xue}
\ead{ycc9gv@virginia.edu}
\author[aff3]{Tien Comlekoglu}
\ead{tc2fh@virginia.edu}
\author[aff4]{Keyi Xue}
\ead{kariexue@live.unc.edu}
\author[aff5]{Guanliang Wang}
\ead{438850070@qq.com}
\author[aff6]{Jiang Li}
\ead{JLi@odu.edu}
\author[aff1,aff2,aff3]{Gordon Laurie \corref{cor1}}
\ead{gwl6s@virginia.edu}
\cortext[cor1]{Corresponding author}

\address[aff1]{Department of Cell Biology, University of Virginia, Charlottesville, VA, USA}
\address[aff2]{Department of Ophthalmology, University of Virginia, Charlottesville, VA, USA}
\address[aff3]{Department of Biomedical Engineering, University of Virginia, Charlottesville, VA, USA}
\address[aff4]{Department of Biostatistics, University of North Carolina at Chapel Hill, Chapel Hill, NC, USA}
\address[aff5]{Department of Oncology, People’s Hospital of Chongqing, Hechuan, Chongqing, China}
\address[aff6]{Department of Electrical and Computer Engineering, Old Dominion University, Norfolk, VA, USA}

% Abstract
\begin{abstract}

\textbf{Purpose:} To establish a foundational performance benchmark for the automated, multi-label classification of eleven retinal diseases, including diabetic retinopathy (DR), glaucoma (GC), and age-related macular degeneration (AMD), using a recently released large synthetic fundus image dataset (SynFundus-1M), and to assess the generalization of this approach on external real-world datasets. 

\textbf{Design:} A retrospective, multi-center deep learning model development and validation study.

\textbf{Methods:} Six deep learning architectures (ConvNeXtV2, SwinV2, ViT, ResNet, EfficientNetV2, and the RETFound foundation model) were trained on the SynFundus-1M dataset, which contains over one million synthetic images generated by a Denoising Diffusion Probabilistic Model. A 5-fold multi-label stratified cross-validation strategy was employed. An XGBoost ensemble model was developed by stacking out-of-fold predictions. The pipeline was validated on three external datasets (a Unified DR Dataset, AIROGS, and RFMiD). Model explainability was investigated using class activation mapping and feature importance analysis.

\textbf{Results:} The ensemble model achieved the highest performance on the internal validation set, with a macro-average Area Under the Receiver Operating Characteristic Curve (AUC) of 0.9973 and an F1-score of 0.9244. On external validation, the ensemble demonstrated strong generalization, achieving an AUC of 0.7972 on a combined DR dataset, 0.9126 on AIROGS, and a macro-AUC of 0.8800 on RFMiD. Pre-training demonstrated superior performance compared to training from scratch. Explainability methods highlighted clinically relevant anatomical features for model predictions.

\textbf{Conclusion:} A deep learning ensemble trained exclusively on synthetic fundus images can accurately classify multiple retinal diseases and generalize effectively to real-world clinical images. This work provides a baseline for future research on large-scale synthetic datasets and establishes synthetic data as a viable pathway for overcoming data scarcity and privacy issues to develop robust, multi-disease AI systems in ophthalmology.
\end{abstract}


\end{frontmatter}

%%%%%%%%%%%%%%%%%%%%%%%
%% Main Body
%%%%%%%%%%%%%%%%%%%%%%%

\section{Introduction}

Retinal diseases, such as diabetic retinopathy (DR), glaucoma, and age-related macular degeneration (AMD), represent a significant global health burden and are leading causes of preventable blindness \cite{flaxman2017global, bourne2017magnitude}. Fundus photography is a cornerstone of ophthalmic screening, enabling the early detection and monitoring of these conditions, which is crucial for timely intervention and preserving vision. However, the increasing prevalence of diseases like diabetes has led to a surge in the volume of retinal screenings, placing a substantial strain on ophthalmologists and trained human graders. This creates a diagnostic bottleneck, limited by high costs, accessibility issues in underserved regions, and the potential for inter-grader variability and fatigue \cite{gulshan2016development, ting2017development}.

Artificial intelligence (AI), particularly deep learning (DL), has emerged as a transformative technology to address these challenges. Convolutional neural networks (CNNs) have demonstrated remarkable success in analyzing medical images, with several algorithms for DR detection achieving diagnostic performance comparable or superior to human experts \cite{gulshan2016development, ting2019artificial}. These successes have paved the way for the development of AI tools aimed at increasing the efficiency, accessibility, and consistency of retinal screening programs.

Despite these advances, the development of robust, large-scale ophthalmic AI faces significant hurdles. The primary limitation is the data bottleneck: acquiring large, diverse, and expertly annotated medical datasets is notoriously difficult due to patient privacy regulations (e.g., HIPAA), the high cost of expert annotation, and the inherent imbalance of disease prevalence \cite{esteva2021deep}. Furthermore, most existing models are designed for single-disease classification (e.g., DR vs. no DR), which does not reflect the clinical reality where multiple pathologies can coexist. A patient screened for DR may have undiagnosed glaucoma or AMD. Therefore, there is a pressing need for multi-label classification systems that can simultaneously detect a wide spectrum of diseases from a single fundus image.

The recent advent of advanced generative models, specifically Denoising Diffusion Probabilistic Models (DDPMs), offers a paradigm-shifting solution to this data scarcity problem \cite{ho2020denoising}. These models can generate highly realistic, novel medical images that are statistically similar to real-world data. The SynFundus-1M dataset, released in 2023 and created using such a model trained on over 1.3 million private clinical images, provides an unprecedented resource of over one million synthetic fundus photographs with rich annotations for eleven distinct retinal diseases \cite{shang2023synfundus}. This allows for the training of complex, data-hungry models without compromising patient privacy. Given its novelty, however, no comprehensive multi-label classification benchmarks currently exist for this dataset. The research community lacks a baseline understanding of how modern architectures perform on this new data modality and, critically, whether models trained on it can generalize to real clinical images.

In this study, we use the SynFundus-1M dataset to address the need for a comprehensive, multi-label retinal disease classifier. To address this clear research gap and establish a foundational benchmark for this major new resource, our study presents the first end-to-end deep learning pipeline for multi-label classification on the SynFundus-1M dataset. We benchmark the performance of six modern deep learning architectures, including CNNs, Vision Transformers (ViT), and a domain-specific foundation model (RETFound). We then develop a meta-ensemble model that integrates the predictive strengths of these individual architectures. Finally, we rigorously evaluate the generalization of our models on three diverse, public, real-world datasets to assess their clinical utility and establish the viability of using large-scale synthetic data for developing next-generation ophthalmic AI.

\section{Methods}
\subsection{Study Design and Dataset}

This retrospective study details the development and validation of a deep learning pipeline for multi-label retinal disease classification using a recent, open-source, publicly available synthetic dataset. All models were subsequently evaluated on multiple external, real-world datasets to assess generalization performance.

The primary dataset used for model development was \textbf{SynFundus-1M}, a large-scale dataset of over one million synthetic fundus images \cite{shang2023synfundus}. The images were generated by a Denoising Diffusion Probabilistic Model (DDPM) that was trained on a private set of 1.3 million authentic clinical fundus photographs. The dataset includes annotations for eleven distinct retinal pathologies, as detailed in Table \ref{tab:diseases}.

\begin{table}[h!]
\centering
\caption{The eleven retinal disease labels targeted for classification.}
\label{tab:diseases}
\begin{tabularx}{0.95\textwidth}{@{}p{3.6cm} p{1.1cm} X@{}}
\toprule
\textbf{Disease} & \textbf{Abbr.} & \textbf{Description} \\
\midrule
Age-related Macular Degeneration & AMD & Degenerative condition affecting the macula, a leading cause of central vision loss. \\
Anomalies of the Optic Nerve & AON & Structural abnormalities of the optic nerve head, such as optic disc drusen or pits. \\
Choroidal Retinal Pathology & CRP & A broad category of pathologies affecting the choroid and retina, including scars and nevi. \\
Degenerative Myopia & DM & Pathological changes in the posterior segment of the eye due to high myopia. \\
Diabetic Macular Edema & DME & Swelling in the macula caused by leaking blood vessels in patients with diabetes. \\
Diabetic Retinopathy & DR & Damage to retinal blood vessels caused by diabetes; grades 1-4 were treated as a single positive class. \\
Epimacular Membrane & EM & A scar-like tissue that forms over the macula, causing blurred and distorted vision. \\
Glaucoma & GC & A group of eye conditions that damage the optic nerve, often associated with high eye pressure. \\
Hypertensive Retinopathy & HtR & Retinal vascular damage caused by chronically high blood pressure. \\
Pathological Myopia & PM & Severe, progressive nearsightedness leading to degenerative changes in the eye. \\
Retinal Vein Occlusion & RVO & A blockage of the small veins that carry blood away from the retina. \\
\bottomrule
\end{tabularx}
\end{table}

Prior to model training, an initial data curation step was performed to verify the existence of all image files referenced in the annotation manifest, ensuring dataset integrity. For the classification task, the original five-level diabetic retinopathy annotation (Grades 0-4) was converted into a binary label representing the presence (Grades 1-4) or absence (Grade 0) of the disease. The complete dataset was then partitioned using a 5-fold cross-validation scheme. To handle the multi-label nature of the data, a \texttt{MultilabelStratifiedKFold} strategy was employed, which preserves the percentage of samples for each of the eleven disease labels in each fold, ensuring a balanced and consistent data distribution for training and validation.

\subsection{Model Architectures}

To provide a comprehensive benchmark of modern deep learning capabilities on this task, six state-of-the-art architectures were selected. The selection included both established and recent models, spanning two primary families of computer vision architectures: Convolutional Neural Networks (CNNs) and Vision Transformers (ViTs). The CNN-based models included ResNet50 \cite{he2016deep}, EfficientNetV2-M \cite{tan2021efficientnetv2}, and ConvNeXtV2-Base \cite{woo2023convnext}. The Transformer-based models were ViT-Base \cite{dosovitskiy2020image}, SwinV2-Base \cite{liu2022swin}, and RETFound \cite{zhou2023foundation}. Models containing a similar number of total parameters were chosen to keep the comparison objective.

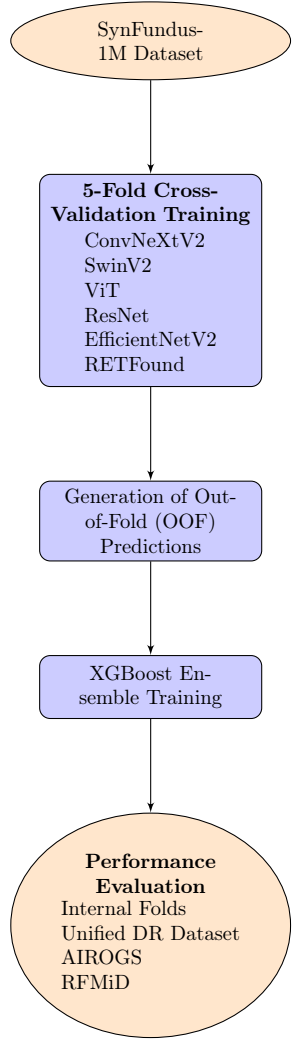
\begin{figure}[htbp]
\centering
\caption{The complete study workflow, from data processing and model training to ensemble construction and final evaluation.}
\label{fig:workflow}
\begin{tikzpicture}[
    scale=0.78,                % overall scale (tweak 0.7--0.9)
    transform shape,           % preserves node proportions when scaling
    node distance=1.6cm,       % less vertical space between nodes
    auto,
    block/.style={
        rectangle,
        draw,
        fill=blue!20,
        text width=8.5em,      % narrower text box
        text centered,
        rounded corners,
        minimum height=2.6em,  % shorter blocks
        font=\footnotesize     % smaller font
    },
    line/.style={draw, -latex'},
    cloud/.style={
        ellipse,
        draw,
        fill=orange!20,
        minimum height=2.6em,
        text width=7.5em,
        text centered,
        font=\footnotesize
    }
]

    % Nodes (use compact tabular instead of itemize)
    \node [cloud] (input) {SynFundus-1M Dataset};

    \node [block, below=of input] (training) {
        \textbf{5-Fold Cross-Validation Training}\\[0.3ex]
        \begin{tabular}{@{}l@{}}
        ConvNeXtV2 \\ SwinV2 \\ ViT \\ ResNet \\ EfficientNetV2 \\ RETFound
        \end{tabular}
    };

    \node [block, below=of training, node distance=3.2cm] (oof) {
        Generation of Out-of-Fold (OOF) Predictions
    };

    \node [block, below=of oof] (ensemble) {XGBoost Ensemble Training};

    \node [cloud, below=of ensemble] (output) {
        \textbf{Performance Evaluation}\\[0.3ex]
        \begin{tabular}{@{}l@{}}
        Internal Folds \\ Unified DR Dataset \\ AIROGS \\ RFMiD
        \end{tabular}
    };

    % Edges
    \path [line] (input) -- (training);
    \path [line] (training) -- (oof);
    \path [line] (oof) -- (ensemble);
    \path [line] (ensemble) -- (output);

\end{tikzpicture}
\end{figure}

All models were implemented in PyTorch \cite{paszke2019pytorch} and sourced from the PyTorch Image Models (\texttt{timm}) library \cite{wightman2021resnet}. RETFound is a foundation model specifically pre-trained on a large dataset of 1.6 million unlabeled retinal images, designed to provide a strong, domain-specific feature extractor for downstream ophthalmic tasks. The complete study workflow, from data processing to final evaluation, is illustrated in Figure \ref{fig:workflow}.

\subsection{Experimental Setup}

\paragraph{Hardware and Software} All model training and inference were conducted on the University of Virginia's Rivanna high-performance computing cluster with NVIDIA A100 GPUs. The deep learning pipeline was developed using the PyTorch framework (version 2.7.0). Data augmentation was implemented using the albumentations library \cite{buslaev2020albumentations}.

\paragraph{Image Preprocessing and Augmentation}
All images were first resized to a standard input resolution of $224 \times 224$ pixels, except for the SwinV2 architecture, which required a $192 \times 192$ input. To improve model robustness and prevent overfitting, a series of data augmentations were applied to the training images. This pipeline included random horizontal flipping (p=0.5), random adjustments to brightness and contrast (p=0.3), random affine transformations including shifting, scaling, and rotation (p=0.4), and elastic transformations (p=0.2). Following augmentation, all images were normalized using the standard mean and standard deviation values from the ImageNet dataset. Validation images were only resized and normalized without further augmentation to ensure a consistent evaluation.

\paragraph{Hyperparameter Optimization}
To determine the optimal training parameters for each of the six architectures, a systematic hyperparameter optimization process was employed using the Optuna framework \cite{akiba2019optuna}. For each model, a 30-trial search was conducted on a 10\% subset of the training data. The search space included three key hyperparameters: the learning rate (log-uniformly sampled between $1 \times 10^{-6}$ and $3 \times 10^{-4}$), weight decay (log-uniformly sampled between $1 \times 10^{-6}$ and $1 \times 10^{-2}$), and the model's dropout rate (uniformly sampled between 0.0 and 0.6). Each trial was run for a maximum of three epochs, and a median pruner was used to terminate unpromising trials early. The objective of the search was to identify the hyperparameter combination that maximized the macro-average AUC on the validation set.

\paragraph{Model Training}
Using the optimized hyperparameters identified for each architecture, the six models were trained on the full dataset using the 5-fold cross-validation scheme described previously. Each model was trained for 10 epochs with a batch size of 32. We used the AdamW optimizer \cite{loshchilov2017decoupled}, a variant of the Adam optimizer with decoupled weight decay. The loss was calculated using the Binary Cross-Entropy with Logits loss function (\texttt{BCEWithLogitsLoss}), which is used for multi-label classification as it evaluates each disease class independently. For each fold, the model checkpoint that achieved the highest macro-average AUC on the corresponding validation set was saved for subsequent evaluation and ensemble construction.

\subsection{Ensemble Meta-learner Model}

To aggregate the diverse predictive capabilities of the six individual architectures, a two-stage meta-ensemble model was constructed using a stacking methodology. This approach trains a second-level model, or meta-learner, to learn the optimal combination of predictions from the first-level base models.

The input features for this meta-learner were the \textbf{out-of-fold (OOF)} predictions generated during the 5-fold cross-validation. For each fold, the five models trained on the training splits were used to make predictions on the held-out validation split. By concatenating the predictions from all five folds, a complete set of predictions for the entire dataset was generated, where each prediction was made by a model that had not seen that particular image during its training. This OOF strategy is crucial to prevent information leakage and create a generalizable meta-dataset. The final feature set for each image consisted of the concatenated probability outputs for all 11 diseases from all six models, resulting in a feature vector of size 66 ($6 \text{ models} \times 11 \text{ diseases}$).

The meta-learner was an \textbf{Extreme Gradient Boosting (XGBoost)} classifier, a highly effective gradient-boosted decision tree algorithm known for its performance and computational efficiency \cite{chen2016xgboost}. To adapt the XGBoost model for the multi-label task, it was wrapped in a \texttt{MultiOutputClassifier} from the scikit-learn library \cite{pedregosa2011scikit}. This wrapper trains a distinct XGBoost classifier for each of the eleven disease labels, allowing the final ensemble to make an independent, optimized prediction for every pathology.

\subsection{External Validation Datasets}

To assess the real-world generalization performance of our models, we evaluated the best-performing single architecture (ConvNeXtV2) and the final XGBoost ensemble on three diverse, publicly available datasets of authentic fundus photographs. Importantly, this evaluation was performed in a \textbf{zero-shot} manner, meaning the models were applied directly to these external datasets without any fine-tuning or re-training. This protocol strictly tests the models' ability to transfer learned knowledge from the synthetic domain to clinical practice.

The three external datasets, summarized in Table \ref{tab:external_datasets}, were chosen to evaluate performance on distinct clinical tasks.

\begin{table}[h!]
\centering
\caption{Summary of external datasets used for model validation.}
\label{tab:external_datasets}
\begin{tabularx}{0.95\textwidth}{@{}l X X@{}}
\toprule
\textbf{Dataset Name} & \textbf{Images Used} & \textbf{Primary Task} \\
\midrule
Unified DR Dataset \cite{diabetic-retinopathy-detection, aptos2019-blindness-detection, decenciere2014feedback} & 9,261 & Binary Diabetic Retinopathy (DR) Classification \\
AIROGS \cite{devente2023airogsartificialintelligencerobust} & 49,119 & Binary Referable Glaucoma (RG) Classification \\
RFMiD \cite{pachade2021retinal} & 620 & Multi-Label Retinal Disease Classification \\
\bottomrule
\end{tabularx}
\end{table}

The diabetic retinopathy validation was performed on the test set of a \textbf{Unified DR Dataset}, which is a combined collection of four public datasets: EyePACS, APTOS 2019, APTOS 2019 (Gaussian Filtered), and Messidor. The original five-level DR grading scale was mapped to a binary task of detecting the presence (grades 1-4) versus absence (grade 0) of diabetic retinopathy. The \textbf{AIROGS} dataset, containing 101,441 images with labels for referable ('RG') versus non-referable glaucoma ('NRG'), was used for glaucoma validation. A subset of 49,119 images was used for the evaluation. For the \textbf{RFMiD} dataset, we used the 620-image test set to evaluate multi-label classification performance.

Since the RFMiD dataset uses a more granular set of 45 disease labels, a specific mapping was established to align its ground truth with our model's outputs. The ground truth for one of our model's classes was considered positive if any of its corresponding labels in the RFMiD dataset were present, as detailed in Table \ref{tab:rfmid_mapping}.

\begin{table}[h!]
\centering
\caption{Mapping of model output labels to the corresponding ground truth labels in the RFMiD dataset.}
\label{tab:rfmid_mapping}
\begin{tabularx}{\textwidth}{@{}l X@{}}
\toprule
\textbf{Model Output Label} & \textbf{Corresponding RFMiD Labels} \\
\midrule
\texttt{dr\_grade} (DR) & DR (Diabetic Retinopathy) \\
\texttt{is\_dme} (DME) & DME (Diabetic Macular Edema) \\
\texttt{is\_amd} (AMD) & ARMD (Age-related Macular Degeneration) \\
\texttt{is\_em} (EM) & ERM (Epiretinal Membrane) \\
\texttt{is\_rvo} (RVO) & BRVO (Branch Retinal Vein Occlusion), CRVO (Central Retinal Vein Occlusion) \\
\texttt{is\_pm} (PM) & MYA (Myopic Atrophy), TSLN (Tessellation), TD (Tilted Disc) \\
\texttt{is\_aon} (AON) & AION (anterior ischemic optic neuropathy,), ODP (optic disc pallor), ODE (optic disc edema), MNF (myelinated nerve fibers), TD (tilted disc), ODPM (optic disc pit maculopathy) \\
\texttt{is\_crp} (CRP) & RS (retinitis), CRS (chorioretinitis), VS (vasculitis), CSR (central serous retinopathy), PT (parafoveal telangiectasia), PTCR (post traumatic choroidal rupture), CF (choroidal folds), RP (retinitis pigmentosa) \\
\bottomrule
\end{tabularx}
\end{table}

\subsection{Ablation Study: Impact of Pre-training}

To quantify the performance contribution of transfer learning, an ablation study was conducted to compare models trained with pre-trained weights against those trained from random initialization (i.e., "from scratch"). Two representative architectures were selected for this study: ConvNeXtV2, representing a model with general-purpose pre-training on the ImageNet dataset, and RETFound, representing a model with domain-specific pre-training on a large corpus of fundus images.

These two models were re-trained on the SynFundus-1M dataset by setting the \texttt{pretrained} parameter to \texttt{False}. All other aspects of the experimental setup, including the 5-fold cross-validation splits, the optimized hyperparameters for each architecture, and the training duration, were kept identical to the main experiments, ensuring that any observed differences in performance can be directly attributed to the impact of the pre-trained weights. The performance of the models trained from scratch was then directly compared against their pre-trained counterparts.

\subsection{Model Explainability and Interpretation}

To confirm that the models' predictions were based on clinically relevant features, and to enhance the transparency of our "black box" models, we employed several explainability techniques. 

\paragraph{Visual Explainability} For individual model architectures, we generated visual heatmaps to highlight the image regions that most influenced the prediction for a specific disease. These methods included gradient-based techniques such as Saliency Maps and Integrated Gradients, which trace the model's output back to input pixel importance. We also used a perturbation-based method, Occlusion Analysis, which systematically blocks portions of the image to measure their impact on the output score. For the CNN-based architectures (ResNet50 and EfficientNetV2-M), we additionally applied Gradient-weighted Class Activation Mapping (Grad-CAM) \cite{selvaraju2017grad} to visualize the feature activations in the final convolutional layer.

\paragraph{Ensemble Feature Importance} To interpret the behavior of the meta-ensemble, we analyzed the internal feature importance scores from the trained XGBoost model. In this context, the "features" are the 66 predictive inputs from the six base models ($6 \text{ models} \times 11 \text{ diseases}$). By examining the importance scores for each of the eleven output classifiers, we could determine the relative contribution of each base architecture to the ensemble's final prediction for every disease class. This provides insight into whether certain models acted as "specialists" for specific pathologies within the ensemble.

\subsection{Statistical Analysis}

Model performance was primarily evaluated using the \textbf{Area Under the Receiver Operating Characteristic Curve (AUC)} and the F1-score. Precision and Recall were also calculated. For the multi-label tasks on the SynFundus-1M and RFMiD datasets, these metrics were first computed on a per-class basis and then macro-averaged. The macro-average computes the metric independently for each class and then takes the average, treating all classes as equally important regardless of their prevalence.

All metric computations were implemented using the \texttt{torchmetrics} and \texttt{scikit-learn} libraries. For the binary classification tasks on the external Unified DR and AIROGS datasets, full Receiver Operating Characteristic (ROC) were generated to visualize model performance across the entire range of decision thresholds.

\section{Results}
\subsection{Hyperparameter Optimization}

Prior to full-scale training, a 30-trial hyperparameter search was conducted for each of the six architectures to identify the optimal learning rate, weight decay, and dropout rate. The combination of parameters that yielded the highest macro-average AUC during the search was selected for the final 5-fold cross-validation training. The optimized hyperparameters for each model are presented in Table \ref{tab:hyperparameters}.

\begin{table}[h!]
\centering
\caption{Optimized hyperparameters for each model architecture identified using the Optuna framework.}
\label{tab:hyperparameters}
\begin{threeparttable}
\begin{tabular}{l 
                S[scientific-notation = true, table-format=1.4e-2, round-precision=2] % Learning Rate
                S[scientific-notation = true, table-format=1.4e-2, round-precision=2] % Weight Decay
                S[scientific-notation = fixed, table-format=1.3, round-precision=3]}  % Dropout Rate
\toprule
\textbf{Architecture} & {\textbf{Learning Rate}} & {\textbf{Weight Decay}} & {\textbf{Dropout Rate}} \\
\midrule
ResNet50 & 2.63e-4 & 8.74e-6 & 0.056 \\
SwinV2-Base & 1.13e-4 & 2.02e-4 & 0.082 \\
EfficientNetV2-M & 2.64e-4 & 1.75e-3 & 0.004 \\
RETFound & 7.91e-5 & 1.76e-4 & 0.055 \\
ViT-Base & 3.92e-5 & 1.00e-6 & 0.208 \\
ConvNeXtV2-Base & 1.29e-4 & 1.04e-5 & 0.010 \\
\bottomrule
\end{tabular}
\end{threeparttable}
\end{table}

\subsection{Performance on SynFundus-1M}

Following the 5-fold cross-validation protocol, all six individual architectures demonstrated a high level of performance in classifying the eleven retinal diseases on the held-out validation sets. The comprehensive results, including the performance of the meta-ensemble, are summarized in Table \ref{tab:cv_performance}. The XGBoost ensemble achieved the highest overall performance, surpassing all individual models in both macro-average AUC and F1-score. Among the single models, ConvNeXtV2-Base showed the strongest results. The low standard deviation in AUC scores across the folds indicates stable and consistent training performance for all architectures.

\begin{table}[h!]
\centering
\caption{5-Fold Cross-Validation Performance on the SynFundus-1M Dataset.}
\label{tab:cv_performance}
\begin{threeparttable}
\sisetup{
    table-format=1.4,
    separate-uncertainty=true
}
\begin{tabular}{l S S[table-format=1.4(1)]}
\toprule
\textbf{Model} & {\textbf{Macro-Avg F1-Score}} & {\textbf{Macro-Avg AUC (Mean $\pm$ SD)}} \\
\midrule
RETFound & 0.9134 & 0.9959(4) \\
ViT-Base & 0.9144 & 0.9958(9) \\
SwinV2-Base & 0.9176 & 0.9964(5) \\
EfficientNetV2-M & 0.9181 & 0.9965(5) \\
ResNet50 & 0.9182 & 0.9965(4) \\
ConvNeXtV2-Base & 0.9186 & 0.9966(6) \\
\midrule
\textbf{XGBoost Ensemble} & \textbf{0.9244} & \textbf{0.9973} \\
\bottomrule
\end{tabular}
\begin{tablenotes}
\item [$\pm$ SD] Mean and sample standard deviation of the best Macro-Average AUC achieved across the 5 validation folds. The F1-Score is the final macro-average calculated on the complete set of out-of-fold predictions. The ensemble AUC is calculated on a 25\% hold-out set of the OOF predictions.
\end{tablenotes}
\end{threeparttable}
\end{table}

% ---- SynFundus multi-label: Confusion matrix + ROC/AUC ----
\begin{figure}[t]
    \centering
    \begin{subfigure}{0.48\textwidth}
        \centering
        \includegraphics[width=\linewidth]{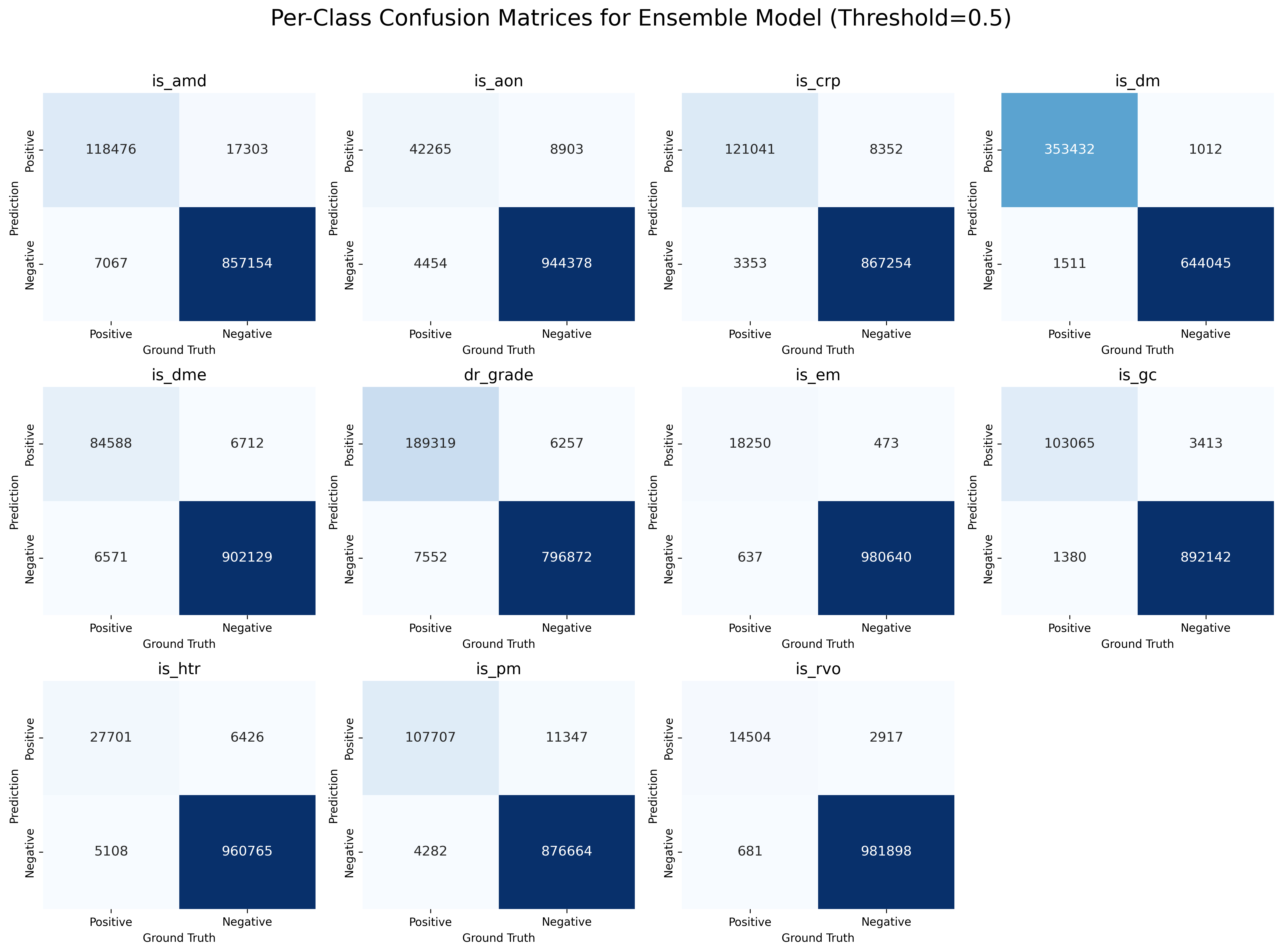}
        \caption{Multi-label confusion matrix (per-class aggregation).}
        \label{fig:synfundus_confmat}
    \end{subfigure}\hfill
    \begin{subfigure}{0.48\textwidth}
        \centering
        \includegraphics[width=\linewidth]{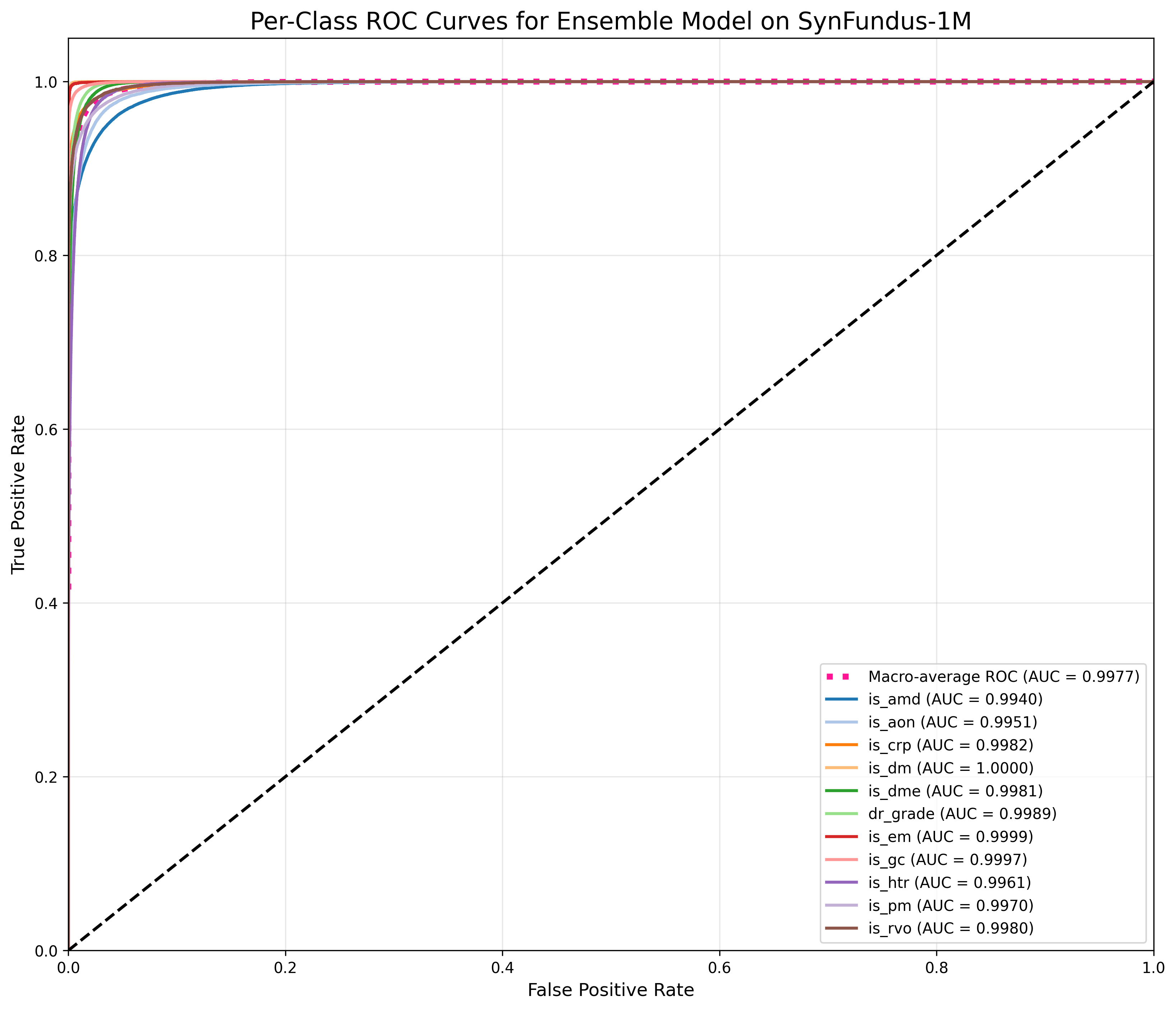}
        \caption{Per-class ROC curves with macro-average AUC.}
        \label{fig:synfundus_roc}
    \end{subfigure}

    \caption{Internal performance on SynFundus-1M. (A) Multi-label confusion matrix summarizing per-class results. (B) Per-class ROC curves and macro-average ROC/AUC.}
    \label{fig:synfundus_confmat_roc}
\end{figure}

A detailed breakdown of the per-class performance for the best single model (ConvNeXtV2-Base) and the XGBoost ensemble is provided in Table \ref{tab:per_class_performance}. The ensemble model demonstrated superior or equal performance compared to the best single model on every disease class for both AUC and F1-score. Both models achieved near-perfect AUC scores ($>$0.999) for several classes, including Degenerative Myopia (DM), Epimacular Membrane (EM), and Glaucoma (GC), while performance was lowest, yet still excellent, on Age-related Macular Degeneration (AMD) and Hypertensive Retinopathy (HtR).

\begin{figure}[H]
    \centering
    \includegraphics[width=\linewidth]{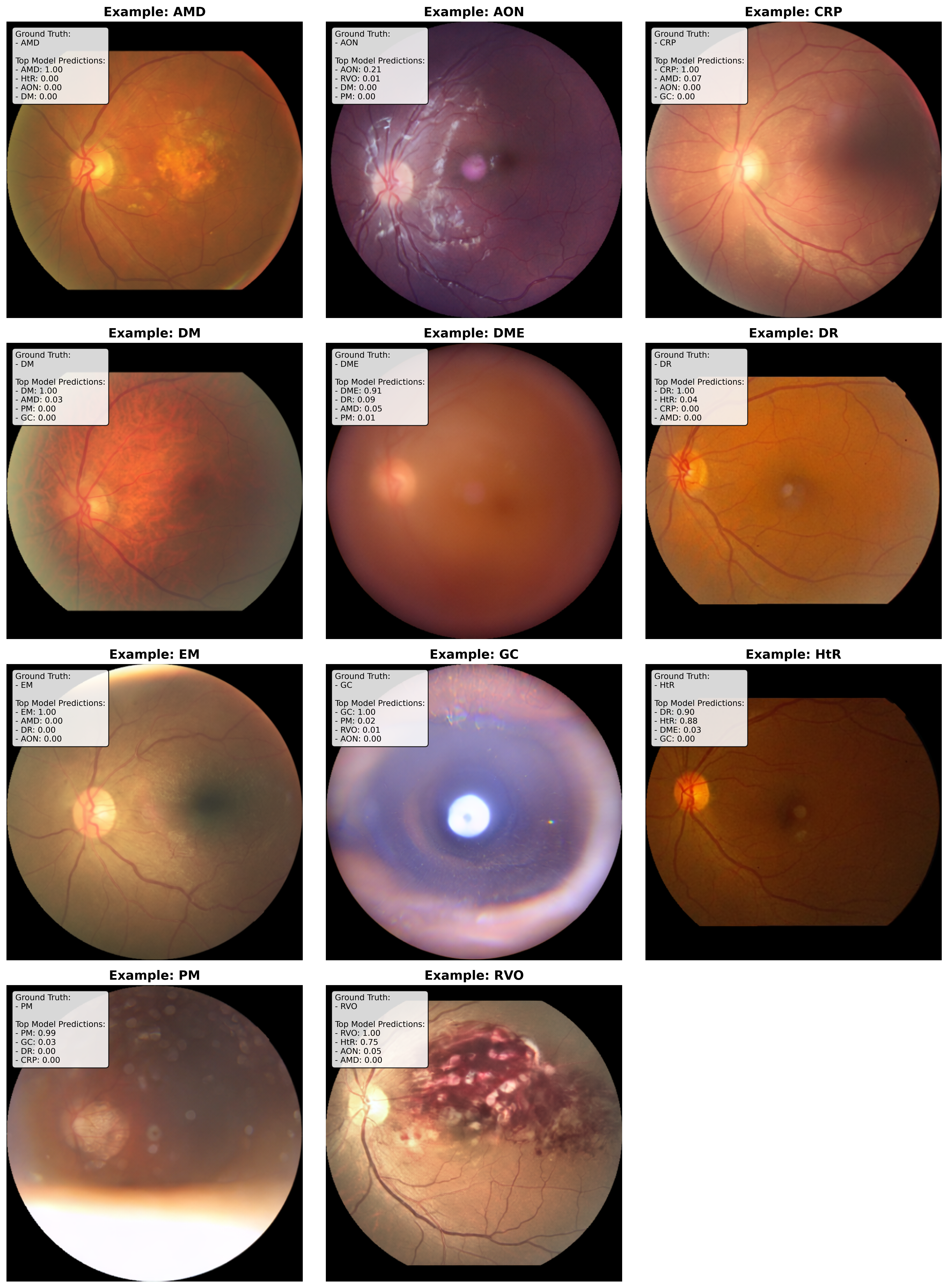}
    \caption{Internal performance on SynFundus-1M. (A) Multi-label confusion matrix summarizing per-class results. (B) Per-class ROC curves and macro-average ROC/AUC.}
    \label{fig:synfundus_predictions}
\end{figure}

\begin{table}[h!]
\centering
\caption{Per-Class Performance of the Best Single Model (ConvNeXtV2) and the Ensemble.}
\label{tab:per_class_performance}
\sisetup{
    table-format=1.4,
    round-mode=places,
    round-precision=4,
    detect-weight
}
\begin{tabular}{l S S S S}
\toprule
& \multicolumn{2}{c}{\textbf{ConvNeXtV2-Base}} & \multicolumn{2}{c}{\textbf{XGBoost Ensemble}} \\
\cmidrule(lr){2-3} \cmidrule(lr){4-5}
\textbf{Disease} & {\textbf{AUC}} & {\textbf{F1-Score}} & {\textbf{AUC}} & {\textbf{F1-Score}} \\
\midrule
AMD & 0.9918 & 0.8957 & \bfseries 0.9933 & \bfseries 0.9043 \\
AON & 0.9928 & 0.8492 & \bfseries 0.9941 & \bfseries 0.8597 \\
CRP & 0.9970 & 0.9469 & \bfseries 0.9979 & \bfseries 0.9526 \\
DM & 0.9999 & 0.9955 & \bfseries 1.0000 & \bfseries 0.9960 \\
DME & 0.9973 & 0.9160 & \bfseries 0.9978 & \bfseries 0.9238 \\
DR & 0.9984 & 0.9569 & \bfseries 0.9988 & \bfseries 0.9627 \\
EM & 0.9996 & 0.9609 & 0.9996 & \bfseries 0.9615 \\
GC & 0.9993 & 0.9731 & \bfseries 0.9996 & \bfseries 0.9758 \\
HtR & 0.9943 & 0.8045 & \bfseries 0.9953 & \bfseries 0.8188 \\
PM & 0.9956 & 0.9251 & \bfseries 0.9966 & \bfseries 0.9302 \\
RVO & 0.9950 & 0.8804 & \bfseries 0.9969 & \bfseries 0.8829 \\
\bottomrule
\end{tabular}
\end{table}

\subsection{Generalization Performance on External Datasets}

To assess the clinical utility of models trained exclusively on synthetic data, we evaluated the best-performing single model (ConvNeXtV2-Base) and the XGBoost ensemble on three distinct, real-world clinical datasets in a zero-shot manner. This process measures the models' ability to generalize from the synthetic training domain to unseen patient data from different sources and for various clinical tasks.

\begin{figure}[h!]
    \centering
    \begin{subfigure}[b]{0.48\textwidth}
        \centering
        \includegraphics[width=\textwidth]{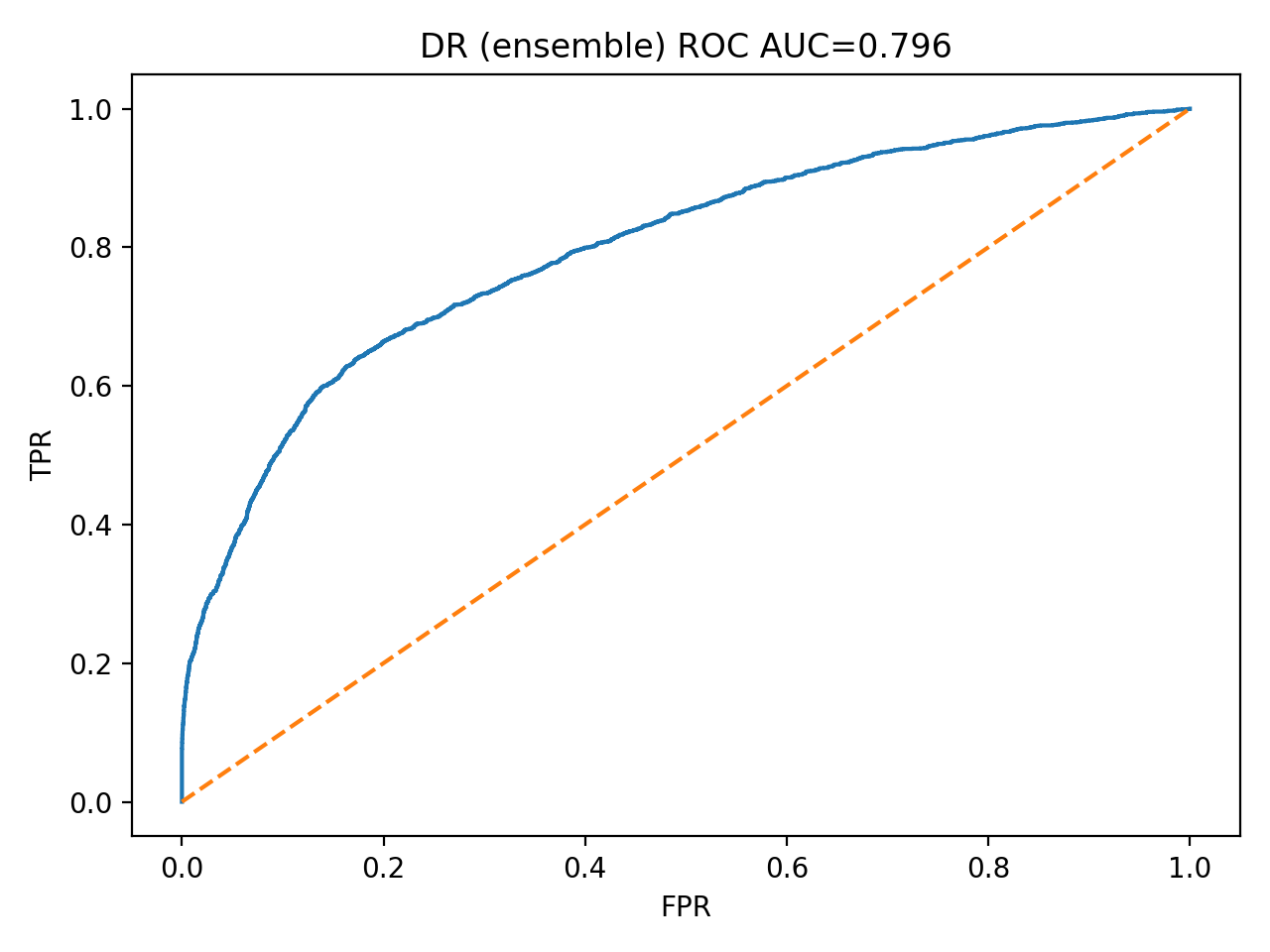} 
        \caption{Ensemble ROC Curve on Unified DR}
        \label{fig:ensemble_roc_dr}
    \end{subfigure}
    \hfill
    \begin{subfigure}[b]{0.48\textwidth}
        \centering
        \includegraphics[width=\textwidth]{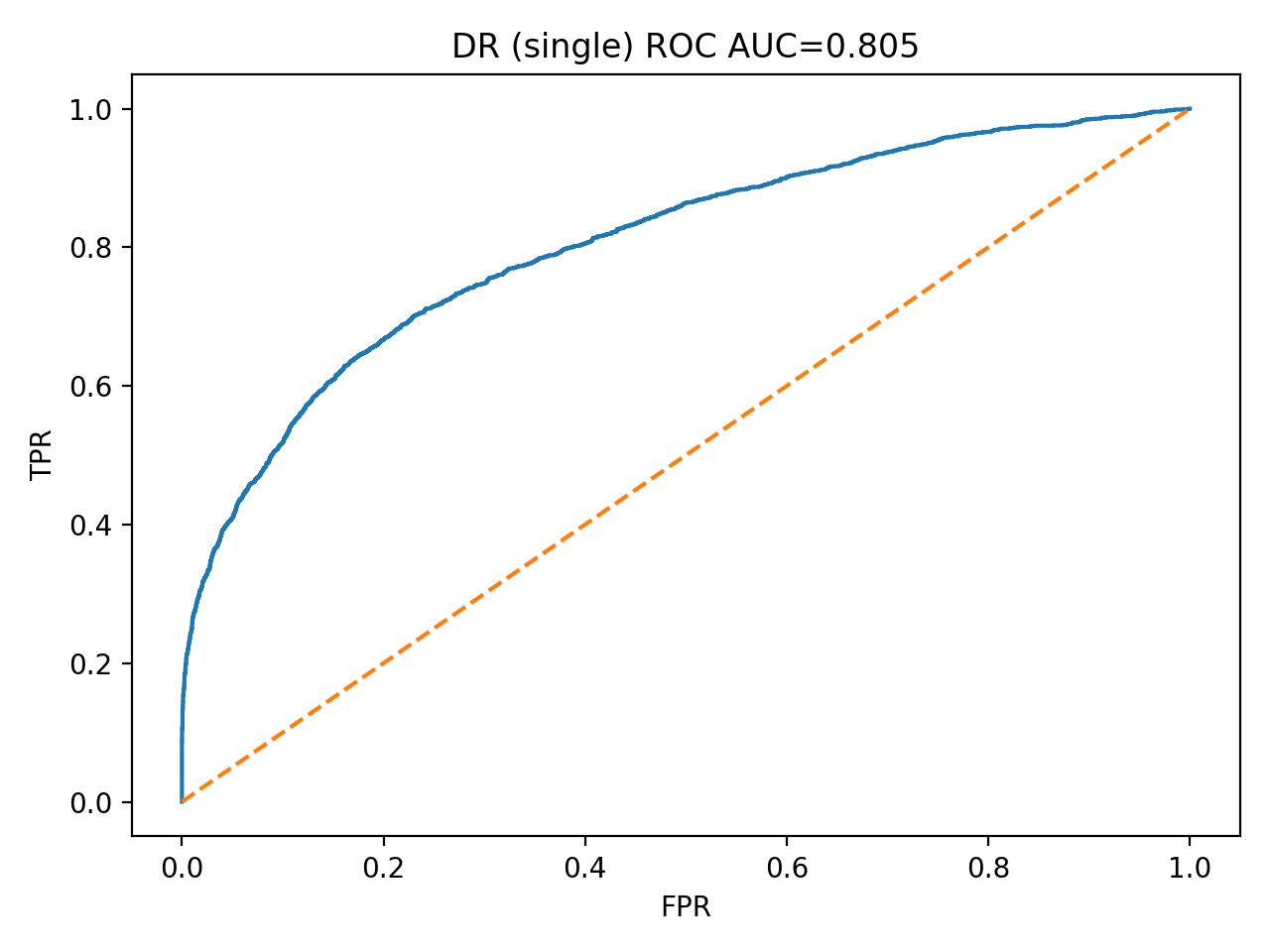} 
        \caption{ConvNeXtV2 ROC Curve on Unified DR}
        \label{fig:convnextv2_roc_dr}
    \end{subfigure}
    \hfill
    
    \vspace{1em} % some vertical space
    
    \begin{subfigure}[b]{0.48\textwidth}
        \centering
        \includegraphics[width=\textwidth]{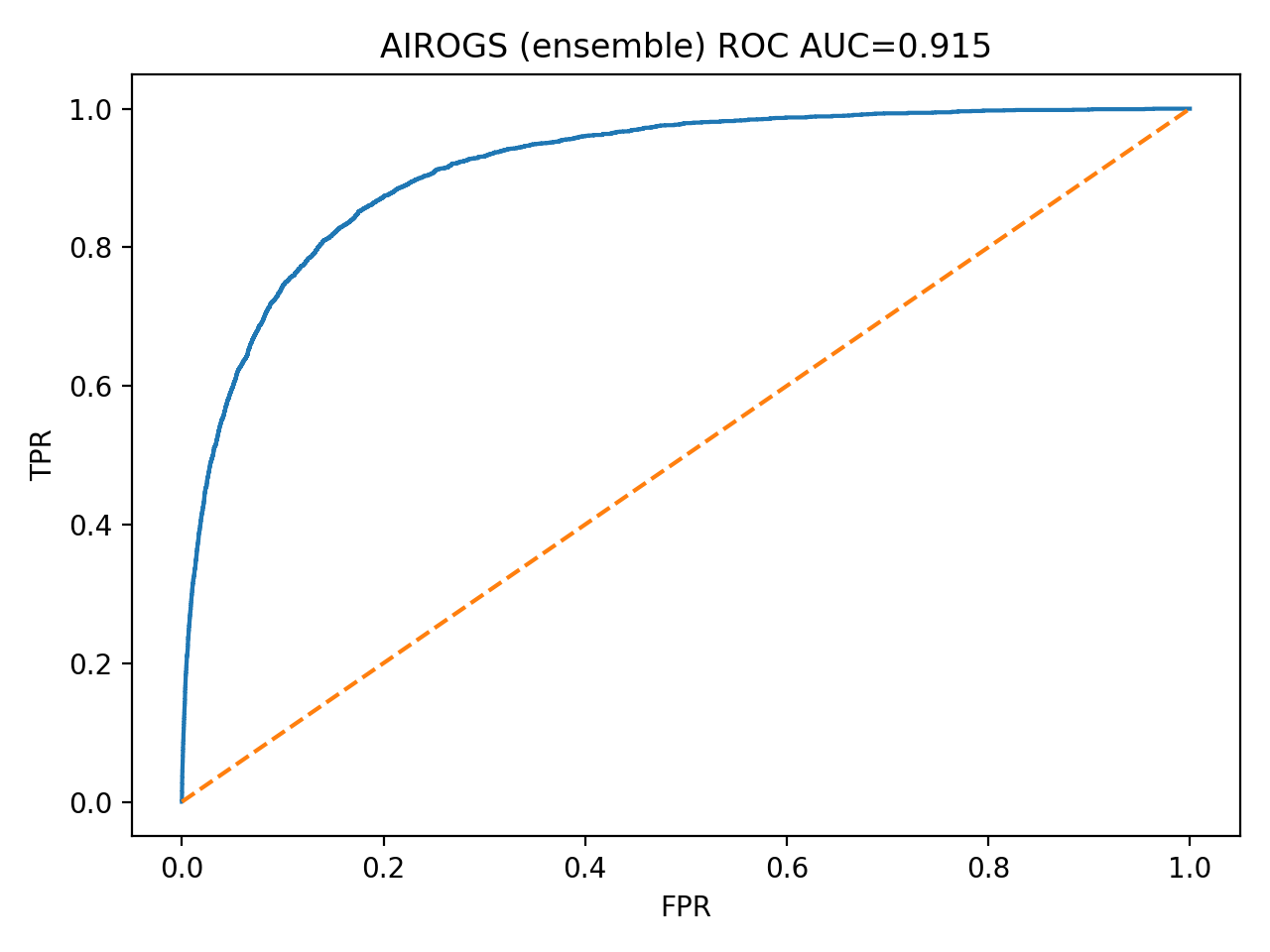}
        \caption{Ensemble ROC Curve on AIROGS}
        \label{fig:ensemble_roc_airogs}
    \end{subfigure}
        \begin{subfigure}[b]{0.48\textwidth}
        \centering
        \includegraphics[width=\textwidth]{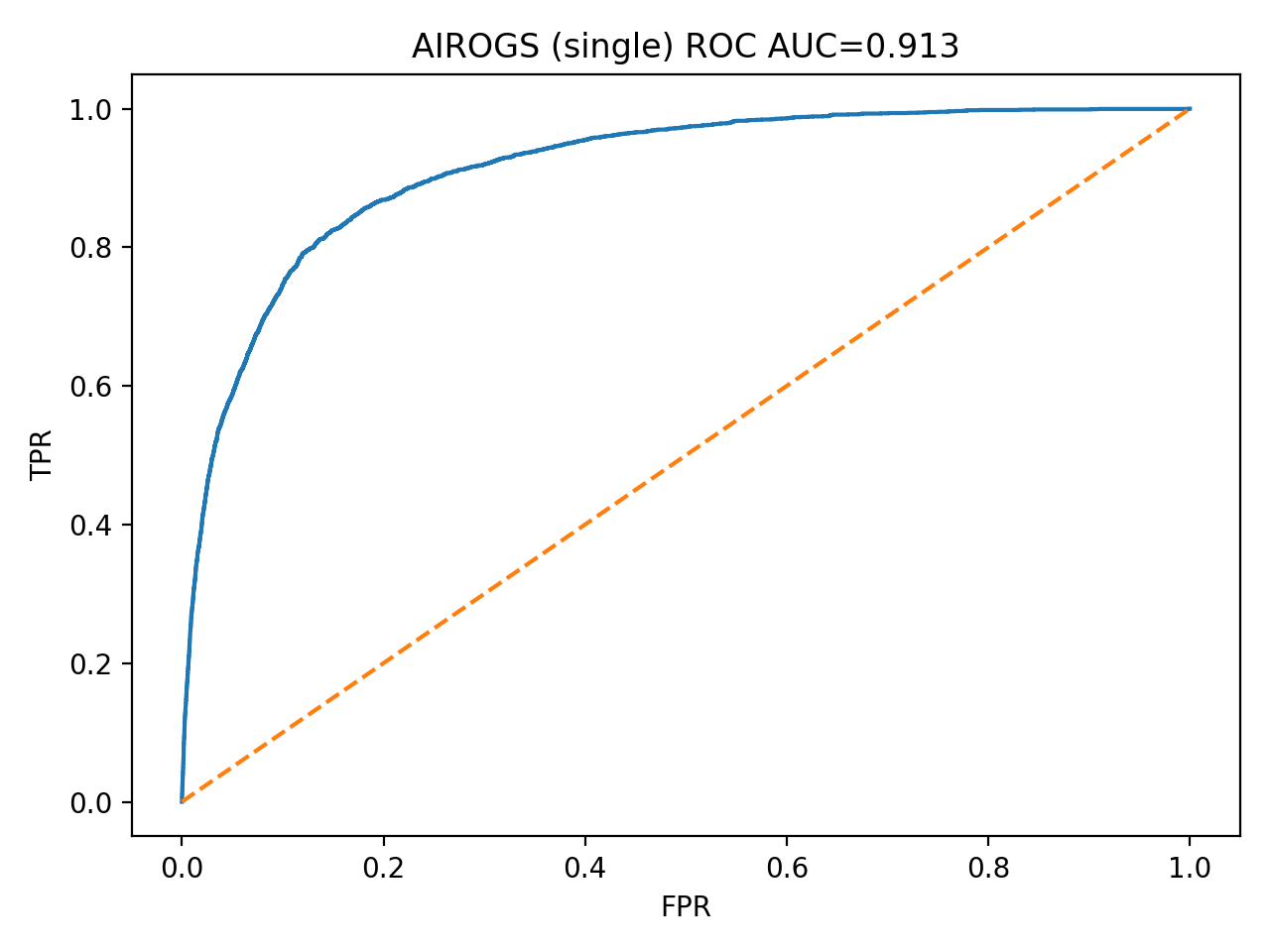}
        \caption{ConvNeXtV2 ROC Curve on AIROGS}
        \label{fig:convnextv2_roc_airogs}
    \end{subfigure}
    \hfill
    \caption{Receiver Operating Characteristic (ROC) curves for the best fold of the single model (ConvNeXtV2-Base) and the XGBoost ensemble on the external validation datasets for (A-B) diabetic retinopathy detection and (C-D) referable glaucoma detection.}
    \label{fig:roc_pr_curves}
\end{figure}

Overall, the models demonstrated strong generalization capabilities. The performance on binary classification tasks for diabetic retinopathy and glaucoma is visualized in Figure \ref{fig:roc_pr_curves}.

A quantitative summary of the Area Under the Curve (AUC) for each task is presented in Table \ref{tab:external_performance}. Both models achieved excellent performance on the AIROGS glaucoma dataset, with the ensemble model reaching an AUC of 0.9126. On the Unified DR Dataset, the models showed respectable performance, with the single ConvNeXtV2 model slightly outperforming the ensemble. For the more complex multi-label RFMiD dataset, the ensemble model demonstrated a clear advantage, achieving a higher macro-average AUC of 0.8800.

\begin{table}[h!]
\centering
\caption{Generalization performance (AUC) on three external validation datasets.}
\label{tab:external_performance}
\sisetup{
    table-format=1.4,
    round-mode=places,
    round-precision=4,
    detect-weight
}
\begin{tabular}{l l S S}
\toprule
\textbf{Dataset} & \textbf{Task} & {\textbf{ConvNeXtV2-Base}} & {\textbf{XGBoost Ensemble}} \\
\midrule
Unified DR & Binary DR & \bfseries 0.8047 & 0.7972 \\
AIROGS & Binary Glaucoma & 0.9091 & \bfseries 0.9126 \\
RFMiD & Multi-Label (Macro) & 0.8711 & \bfseries 0.8800 \\
\bottomrule
\end{tabular}
\end{table}

These results validate the primary hypothesis of the study, confirming that models trained entirely on synthetic data can successfully transfer their learned knowledge to perform a variety of classification tasks on authentic clinical images with high accuracy.

\subsection{Ablation Study: Impact of Pre-training}

An ablation study was conducted to quantify the impact of transfer learning by comparing the performance of models initialized with pre-trained weights against those trained from random initialization (from scratch). The results, shown in Table \ref{tab:ablation_study}, indicate a clear and consistent performance benefit when using pre-trained weights.

For the ConvNeXtV2-Base model, pre-training on ImageNet resulted in a mean macro-average AUC of 0.9967, compared to 0.9951 when trained from scratch. The performance delta was even more pronounced for the domain-specific RETFound model, which achieved a mean AUC of 0.9959 with pre-training versus 0.9924 without. This finding underscores the value of both general-purpose and, particularly, domain-specific pre-training, as it provides a superior feature-learning starting point even when fine-tuning on a dataset as large as SynFundus-1M.

\begin{table}[h!]
\centering
\caption{Ablation study comparing the 5-fold average performance of models with and without pre-trained weights.}
\label{tab:ablation_study}
\sisetup{
    table-format=1.4,
    round-mode=places,
    round-precision=4
}
\begin{tabular}{l l S}
\toprule
\textbf{Model} & \textbf{Training Strategy} & {\textbf{Macro-Average AUC}} \\
\midrule
\multirow{2}{*}{ConvNeXtV2-Base} & Pre-trained (ImageNet) & 0.9967 \\
 & From Scratch & 0.9951 \\
\midrule
\multirow{2}{*}{RETFound} & Pre-trained (Retinal Images) & 0.9959 \\
 & From Scratch & 0.9924 \\
\bottomrule
\end{tabular}
\end{table}

\subsection{Model and Ensemble Explainability}

To ensure the models learned clinically relevant patterns rather than spurious correlations, we investigated their decision-making processes using visual explainability methods and an analysis of the ensemble's feature importances. 

\begin{figure}[t]
    \centering

    % Row 1 (A–D)
    \begin{subfigure}{0.22\textwidth}
        \includegraphics[width=\linewidth]{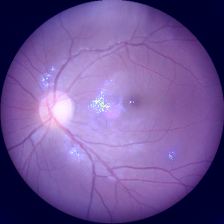}
        \caption{SynFundus-1M}
        \label{fig:saliency_A}
    \end{subfigure}\hfill
    \begin{subfigure}{0.22\textwidth}
        \includegraphics[width=\linewidth]{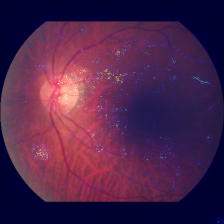}
        \caption{SynFundus-1M}
        \label{fig:saliency_B}
    \end{subfigure}\hfill
    \begin{subfigure}{0.22\textwidth}
        \includegraphics[width=\linewidth]{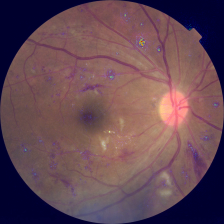}
        \caption{Unified DR}
        \label{fig:saliency_C}
    \end{subfigure}\hfill
    \begin{subfigure}{0.22\textwidth}
        \includegraphics[width=\linewidth]{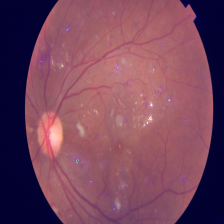}
        \caption{Unified DR}
        \label{fig:saliency_D}
    \end{subfigure}

    \vspace{0.5\baselineskip}

    % Row 2 (E–H)
    \begin{subfigure}{0.22\textwidth}
        \includegraphics[width=\linewidth]{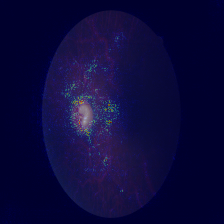}
        \caption{AIROGS}
        \label{fig:saliency_E}
    \end{subfigure}\hfill
    \begin{subfigure}{0.22\textwidth}
        \includegraphics[width=\linewidth]{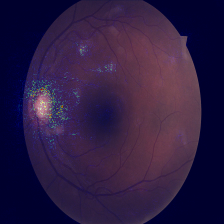}
        \caption{AIROGS}
        \label{fig:saliency_F}
    \end{subfigure}\hfill
    \begin{subfigure}{0.22\textwidth}
        \includegraphics[width=\linewidth]{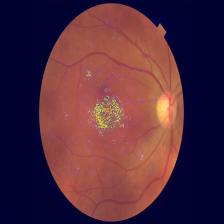}
        \caption{RFMiD}
        \label{fig:saliency_G}
    \end{subfigure}\hfill
    \begin{subfigure}{0.22\textwidth}
        \includegraphics[width=\linewidth]{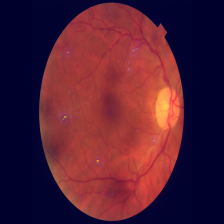}
        \caption{RFMiD}
        \label{fig:saliency_H}
    \end{subfigure}

    \caption{Visual explainability of model predictions. Heatmaps (e.g., Integrated Gradients) are overlaid on fundus images, highlighting regions of high importance. (A–B) Two examples from SynFundus-1M. (C–D) Two from the Unified DR dataset. (E–F) Two from AIROGS. (G–H) Two from RFMiD.}
    \label{fig:saliency_maps}
\end{figure}

\paragraph{Visual Explainability} Saliency and gradient-based attribution methods were used to generate heatmaps that highlight the pixels most influential to a model's prediction for a specific disease. As shown in representative cases in Figure \ref{fig:saliency_maps}, the models consistently focused on clinically appropriate regions. For diabetic retinopathy, the models localized areas with microaneurysms and hemorrhages. For glaucoma, attention was concentrated on the optic nerve head and surrounding nerve fiber layer, which are critical for diagnosis. This provides evidence that the models, despite being trained on synthetic data, learned to identify genuine pathological features.

% ---- Visual explainability grid (8 panels, 2 rows) ----

\begin{figure}[h!]
    \centering
    \begin{subfigure}[b]{0.8\textwidth}
        \centering
        \includegraphics[width=\textwidth]{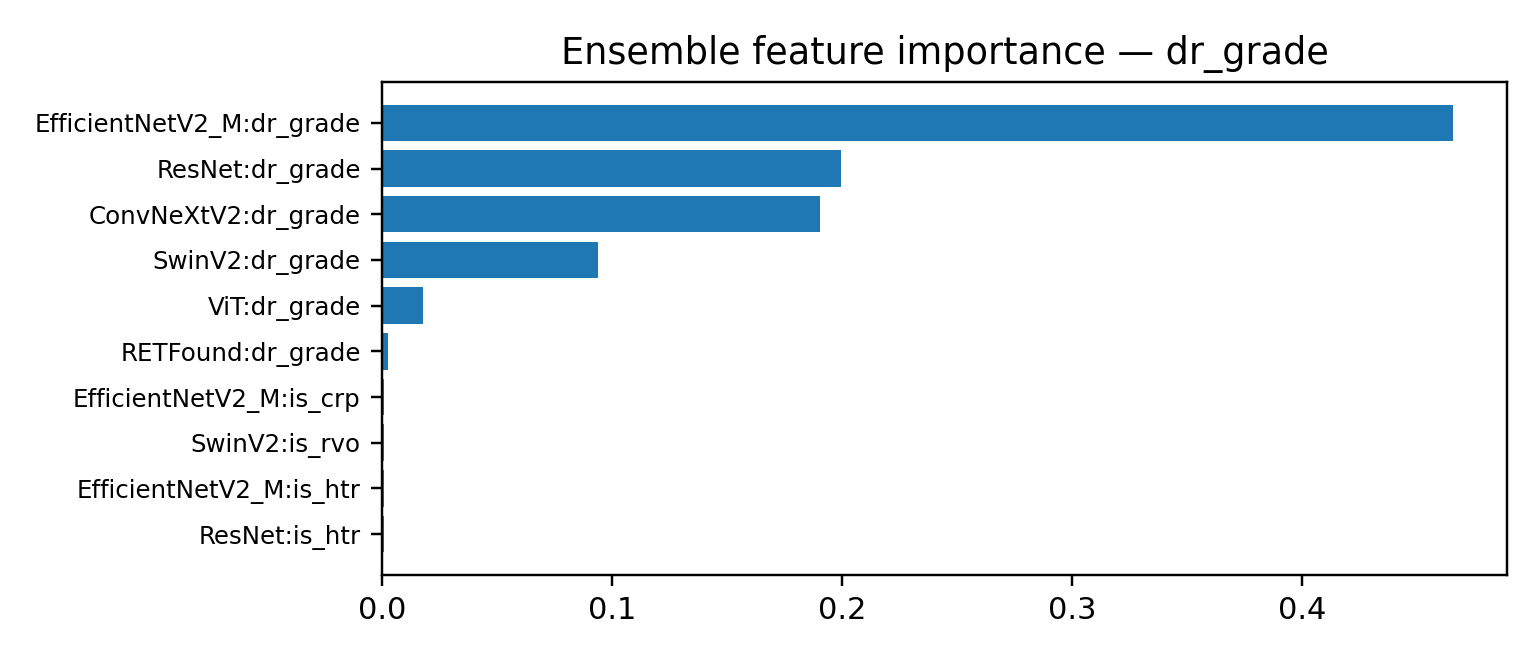}
        \caption{Feature importance for Diabetic Retinopathy (DR)}
        \label{fig:fi_dr}
    \end{subfigure}
    
    \vspace{1em}
    
    \begin{subfigure}[b]{0.8\textwidth}
        \centering
        \includegraphics[width=\textwidth]{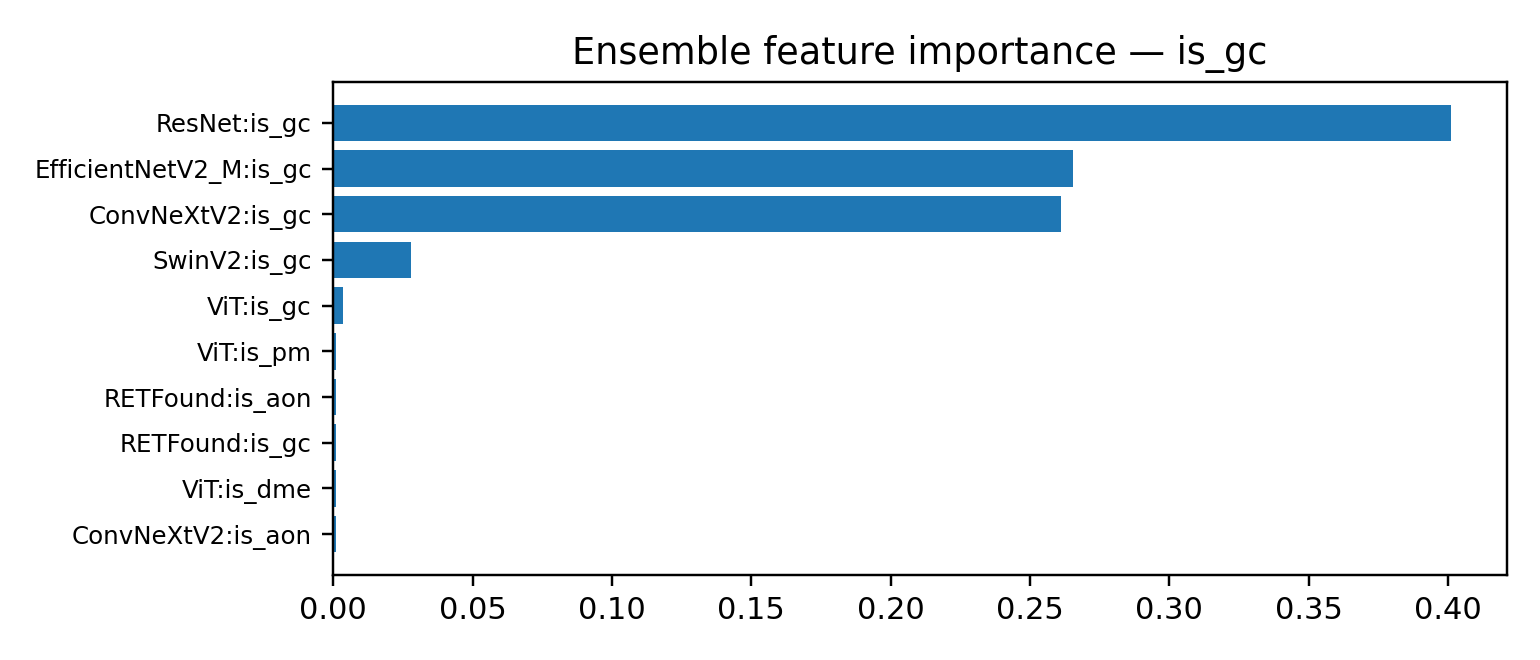}
        \caption{Feature importance for Glaucoma (GC)}
        \label{fig:fi_gc}
    \end{subfigure}

    \vspace{1em}

    \begin{subfigure}[b]{0.8\textwidth}
        \centering
        \includegraphics[width=\textwidth]{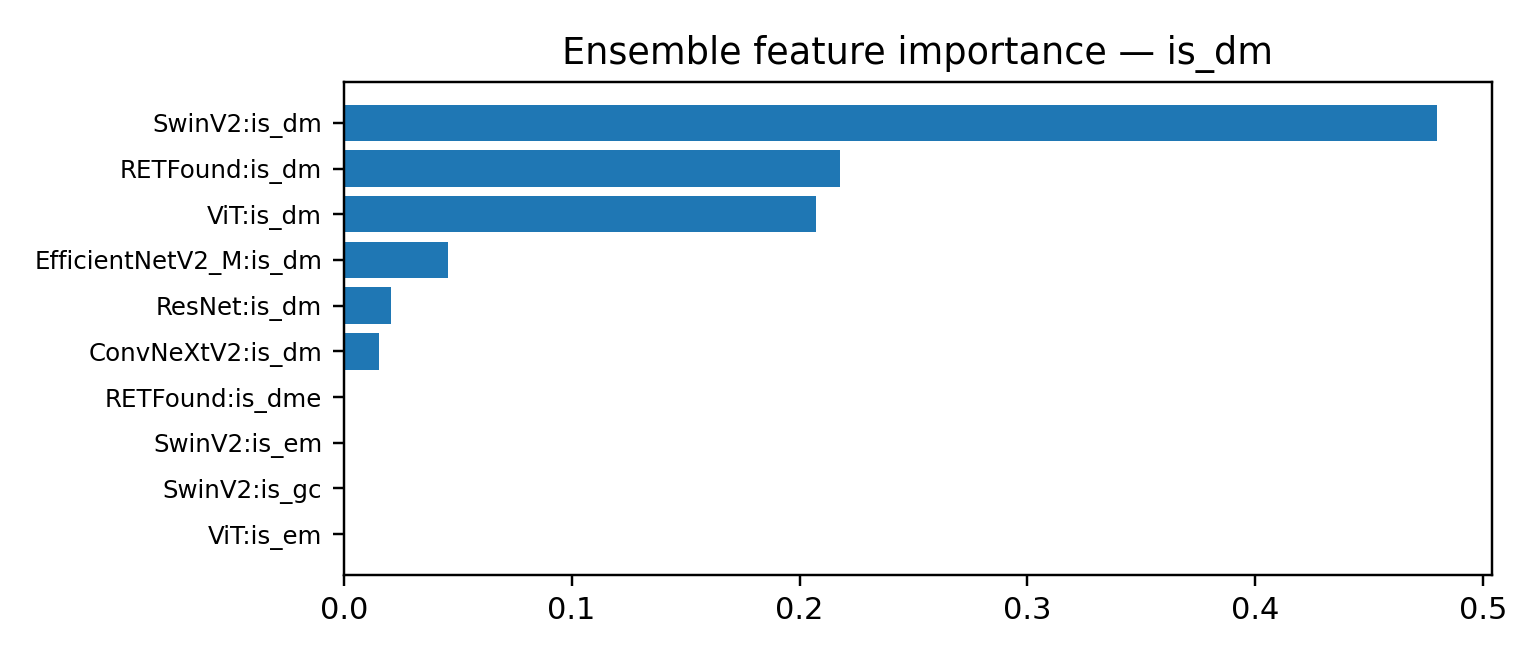}
        \caption{Feature importance for Degenerative Myopia (DM)}
        \label{fig:fi_dm}
    \end{subfigure}    
    \caption{Top 10 most important features for the XGBoost ensemble's predictions for (A) Diabetic Retinopathy, (B) Glaucoma, and (C) Degenerative Myopia. Features are named as `ModelName:DiseaseClass`, indicating the prediction output from each base model.}
    \label{fig:feature_importance}
\end{figure}

\paragraph{Ensemble Feature Importance} Analysis of the XGBoost meta-learner revealed that the ensemble leveraged different base models for different diseases (Figure \ref{fig:feature_importance}). For example, for diagnosing diabetic retinopathy, which relies on detecting fine-grained lesions, the predictions from the convolutional architectures (EfficientNetV2, ResNet) were most influential. Conversely, for degenerative myopia detection, where structural changes are important, the ensemble weighed predictions from the Transformer-based models (SwinV2, RETFound, ViT) more heavily. This indicates that the ensemble learned to act as a committee of specialists, assigning more weight to the model best suited for each pathology.

\section{Discussion}

In this study, we demonstrated that a deep learning pipeline trained exclusively on a large-scale synthetic dataset can accurately and robustly classify eleven distinct retinal diseases. Our primary finding is that a meta-ensemble of modern architectures not only achieved exceptional performance on the internal validation set (Macro-AUC 0.9973) but also generalized remarkably well in a zero-shot setting to three diverse, real-world clinical datasets. This successful transfer of knowledge from a synthetic to a real domain validates our central hypothesis. Furthermore, our results quantitatively confirmed the significant performance benefit of transfer learning, with domain-specific pre-training (RETFound) and general-purpose pre-training (e.g., ImageNet for ConvNeXtV2) both outperforming models trained from scratch.

The ability to develop a high-performing, multi-label classifier without direct access to protected patient health information represents a significant step forward for ophthalmic AI. Traditionally, developing such models has been hampered by the logistical, financial, and ethical challenges of curating large, expertly annotated clinical datasets \cite{ting2019artificial, esteva2021deep}. Our work suggests that high-fidelity synthetic data can serve as an effective proxy, drastically lowering the barrier to entry for developing complex diagnostic models. The performance of our synthetically-trained ensemble on external datasets, such as achieving an AUC of 0.9126 for referable glaucoma detection on AIROGS, is competitive with models specifically trained on real clinical data for that single task \cite{devente2023airogsartificialintelligencerobust}. This indicates that the features learned from the synthetic domain are robust and clinically relevant, a conclusion further supported by explainability analyses that showed our models focused on appropriate anatomical structures.

This study has several strengths. First, the use of the massive SynFundus-1M dataset allowed us to train and benchmark multiple state-of-the-art architectures comprehensively. Second, our methodology was rigorous, incorporating a 5-fold multi-label stratified cross-validation, systematic hyperparameter optimization, and the development of a superior-performing meta-ensemble. Third, and most importantly, the zero-shot validation on three separate external datasets provides strong evidence of the models' generalization capabilities. However, we must also acknowledge the study's limitations. The SynFundus-1M dataset, while realistic, may produce exaggerated or "typical" disease manifestations, potentially limiting the models' sensitivity to subtle or atypical pathologies \cite{shang2023synfundus}. The labels themselves were AI-generated and may contain a degree of noise. Finally, this was a retrospective validation on static datasets; performance in a prospective clinical workflow remains to be evaluated.

Future work should focus on bridging the remaining gap between the synthetic and clinical domains. A promising direction would be to use the models developed in this study as a powerful pre-trained foundation, which could then be fine-tuned on smaller, expert-labeled sets of real clinical data to refine their performance and adapt them to specific patient populations or imaging hardware. Prospective clinical trials are the necessary next step to validate the real-world utility and safety of these models as decision support tools. Ultimately, this work establishes the use of large-scale synthetic datasets as a viable and powerful paradigm for accelerating the development of comprehensive, generalizable, and clinically relevant AI tools in ophthalmology.

\section{Conclusion}

In this study, we developed and benchmarked a deep learning pipeline, culminating in a high-performance meta-ensemble, for the automated multi-label classification of eleven retinal diseases. Crucially, all models were trained exclusively on a large-scale synthetic fundus image dataset, demonstrating a novel paradigm that circumvents traditional barriers of clinical data scarcity and patient privacy. Our synthetically-trained ensemble not only achieved outstanding accuracy on internal validation (Macro-AUC 0.9973) but also demonstrated robust generalization, accurately identifying pathologies in three diverse, unseen external datasets of real-world clinical images. This work establishes that leveraging large-scale, high-fidelity synthetic data is a viable and powerful strategy for building comprehensive and clinically applicable AI models, accelerating the development of next-generation diagnostic tools in ophthalmology.

%%%%%%%%%%%%%%%%%%%%%%%
%% Acknowledgements
%%%%%%%%%%%%%%%%%%%%%%%

\section*{Acknowledgements and Financial Disclosures}
\textbf{a. Funding/Support:} This work was supported in part by the National Institutes of Health under grants 2R01 EY026171 and R01 EY032956.  The authors thank Dr. Fangxin Shang and the authors of the SynFundus-1M paper for their generosity in providing access to their dataset for this research as well as the maintainers of EYEPACs, AIROGS, RFMiD, APTOS, and Messidor.\\
\textbf{b. Financial Disclosures:} No financial disclosures.\\
\textbf{c. Other Acknowledgments:} The authors thank the University of Virginia’s Research Computing group for providing access to the Rivanna HPC cluster. Special acknowledgment is extended to Marcus Bobar and the UVA Research Computing Staff for their technical assistance in resolving dependency conflicts.

%%%%%%%%%%%%%%%%%%%%%%%
%% References
%%%%%%%%%%%%%%%%%%%%%%%

\bibliographystyle{elsarticle-num}
\bibliography{refs}

\appendix

\end{document}